%% file: main.tex
\documentclass[12pt]{article}
\usepackage{arxiv}

\usepackage{graphicx}
\usepackage{subcaption}
\usepackage{amsmath}
\usepackage{xcolor}
\usepackage{multirow}
\usepackage{pifont}
\usepackage{xspace}
\usepackage{comment}
\usepackage[inline]{enumitem}
\usepackage{stfloats}
\usepackage[pagebackref,breaklinks,colorlinks]{hyperref}

\newcommand{\tdevice}{Meta Quest Pro\xspace}
\newcommand{\tdataset}{VRBiom\xspace}
\newcommand{\tbf}{\textit{bona-fide}\xspace}
\newcommand{\cmark}{\textcolor{green}{\ding{51}}}%
\newcommand{\xmark}{\textcolor{red}{\ding{55}}}%

\title{\tdataset: A New Periocular Dataset for Biometric Applications of HMD} 

\author{
Ketan Kotwal$^{1}$,
Ibrahim Ulucan$^{1}$,
G\"{o}khan \"{O}zbulak$^{1,2}$,\\
\vspace{5pt}
\textbf{Janani Selliah$^{1}$, and
S\'{e}bastien~Marcel$^{1,3}$}\\
$^1$Idiap Research Institute, Switzerland\\
$^2$\'{E}cole Polytechnique F\'{e}d\'{e}rale de Lausanne, Switzerland\\
$^3$Universit\'{e} de Lausanne, Switzerland\\
\texttt{\{firstname.lastname\}@idiap.ch}\\%
} 
\date{}

\hypersetup{ 
pdftitle={\tdataset: A New Periocular Dataset for Biometric Applications of HMD}, 
pdfsubject={\tdataset},
pdfauthor={Ketan Kotwal}, 
pdfkeywords={Virtual Reality Dataset, Biometrics, Iris Recognition, Periocular 
Recognition, Presentation Attack Detection, Semantic Segmentation.},
} 
\begin{document}

\maketitle 
 
\begin{abstract}
\input{sections/abstract}
\end{abstract}


\section{Introduction} 
\label{sec:intro} 
\input{sections/intro} 
%
\section{Related Datasets} 
\label{sec:related_work} 
\input{sections/related_work} 
%
\section{The \tdataset Dataset} 
\label{sec:dataset} 
\input{sections/dataset}

%
\section{Potential Use-Cases of the \tdataset Dataset} 
\label{sec:use_cases}
\input{sections/use_cases}

%
\section{Summary} 
\label{sec:conc} 
\input{sections/conc} 
%
\section*{Acknowledgements}
Authors would like to thank Yannick Dayer (Idiap Research Institute) for
creating a data capture application. We would like to gratefully acknowledge
funding support from the Meta program ``Towards Trustworthy Products in AR, VR,
and Smart Devices''
(2022)\footnote{\href{https://research.facebook.com/research-awards/2022-towards-trustworthy-products-in-ar-vr-and-smart-devices}{Meta
Research Program}} and the Swiss Center for Biometrics Research and Testing.

\bibliographystyle{IEEEbib} 
\bibliography{refs} 
 
\end{document}

%% file: sections/abstract.tex

With advancements in hardware, high-quality head-mounted display (HMD) devices
are being developed by numerous companies, driving increased consumer interest
in AR, VR, and MR applications. This proliferation of HMD devices opens up
possibilities of wide range of applications beyond entertainment. Most
commercially available HMD devices are equipped with internal inward-facing
cameras to record the periocular areas. Given the nature of these
devices and captured data, many applications such as biometric authentication
and gaze analysis become feasible. To effectively explore the potential of HMDs
for these diverse use-cases and to enhance the corresponding techniques, it is
essential to have an HMD dataset that captures realistic scenarios.

In this work, we present a new dataset, called \textbf{VRBiom}, of periocular
videos acquired using a Virtual Reality headset. The \tdataset, targeted at
biometric applications, consists of 900 short videos acquired from 25
individuals recorded in the NIR spectrum. These 10$s$ long videos have been
captured using the internal tracking cameras of \tdevice at 72 FPS. To encompass
real-world variations, the dataset includes recordings under three gaze
conditions: steady, moving, and partially closed eyes. We have also ensured an
equal split of recordings without and with glasses to facilitate the analysis of
eye-wear. These videos, characterized by non-frontal views of the eye and
relatively low spatial resolutions ($400 \times 400$), can be instrumental in
advancing state-of-the-art research across various biometric applications. The
\tdataset dataset can be utilized to evaluate, train, or adapt models for
biometric use-cases such as iris and/or periocular recognition and associated
sub-tasks such as detection and semantic segmentation.

In addition to data from real individuals, we have included around 1100
presentation attacks constructed from 92 PA instruments. These PAIs fall into
six categories constructed through combinations of print attacks (real and
synthetic identities), fake 3D eyeballs, plastic eyes, and various types of
masks and mannequins. These PA videos, combined with genuine (\tbf) data, can be
utilized to address concerns related to \textit{spoofing}, which is a
significant threat if these devices are to be used for authentication.

The \tdataset dataset is publicly available for research purposes related to
biometric applications only.

%% file: sections/intro.tex

The rise of head-mounted displays (HMDs) in recent years has significantly
transformed the way we experience digital content. With its Meta Quest series of
virtual reality (VR) headsets, the Meta (previously FaceBook) is the largest
headset platform as of 2024\footnote{\url{https://www.meta.com/quest}}.
Recently, Apple released its Mixed reality (MR) headset named Apple Vision
Pro\footnote{\url{https://www.apple.com/apple-vision-pro}}, with specific focus
on the spatial computing aspects of the device. The Sony Interactive
Entertainment has been actively working on VR for gaming applications with its
PlayStation (PS VR)
series\footnote{\url{https://www.playstation.com/en-us/ps-vr2}}. Initially,
these devices were primarily designed for immersive entertainment, offering
users experiences in augmented, virtual, or mixed reality. However, the
applications of HMDs extend far beyond gaming and entertainment, encompassing
diverse fields such as education~\cite{radianti2020systematic,
kaminska2019virtual}, professional training~\cite{xie2021review,
chiang2022augmented}, healthcare~\cite{yeung2021virtual, javaid2020virtual,
suh2023current}, and biometric authentication~\cite{boutros2020iris,
boutros2020benchmarking}.

Besides its core functionality, an interesting feature of HMDs is the integrated
multiple internal cameras. These cameras, typically located in the surroundings
of the user's eye region, enhance the immersive experience by tracking user's
eye movements and capturing images and videos of the user's eyes and surrounding
regions. This source of data opens up a wide range of potential applications to
be explored, including biometric authentication. In biometrics, the HMDs can be
instrumental in user recognition based on their iris and/or periocular
traits~\cite{boutros2020benchmarking}. Such applications can be employed for
both identification and verification purposes ensuring that the user accessing
the device is indeed who they claim to be. In addition to the recognition
accuracy, in biometric applications, safeguarding against attacks and ensuring
robustness is critical. The presentation attacks (PA), also known as
\textit{spoofing}, pose a serious challenge to biometric systems. These attacks
can involve the use of masks, synthetic eyes, contact lenses, or printed images
to deceive the system into granting unauthorized
access~\cite{morales2019introduction, czajka2018presentation}. Therefore,
developing effective presentation attack detection (PAD) or
\textit{anti-spoofing} mechanisms is essential for the practical use of
biometric authentication systems in HMDs. Beyond biometrics, HMDs have
significant potential in other applications, such as semantic segmentation of
the eye region~\cite{boutros2020iris, wang2021edge}. This is particularly
relevant for industrial, entertainment, and biomedical applications where
continuous tracking of the gaze and eye movement is necessary. HMDs can
facilitate near-continuous gaze tracking, which has applications ranging from
interactive gaming to medical diagnostics~\cite{clay2019eye, kapp2021arett,
bozkir2023eye}. Fig.~\ref{fig:use_cases} depicts some of the possible biometric
use-cases of the HMD data.

\input{sections/fig_usecase_collection}%

The aforementioned applications have been a part of research and commercial
deployments for several years. However, most existing research and
user-applications have focused on data captured by sensors positioned to capture
a frontal view of the eye. The design of recent popular HMDs typically places
displays in front of the user's eyes for an immersive experience, and thereby
necessitates the eye tracking cameras to be positioned in the surrounding areas.
This placement results in an oblique, non-frontal view of the eye
region, posing unique challenges for biometric applications. Additionally, the
cameras integrated into these HMDs are often of lower resolution and must
operate within constrained resources. Moreover, the fitment of HMDs varies based
on the individual's inter-eye distance and nose bridge shape, resulting in
variations in the captured regions for different users.

These factors indicate the need for specific studies and datasets acquired
directly from HMD devices to address the above challenges. Such datasets can be
invaluable for multiple purposes:
\begin{enumerate*}[label=(\textit{\roman*})]
\item benchmarking the performance of existing methods/models on HMD data,
\item fine-tuning models to address domain shifts (introduced by different
capturing angle, devices, resolutions, etc), and 
\item developing new methods tailored to the unique requirements of HMDs.
\end{enumerate*}

To address these challenges, in the realm of biometrics, we have created
\textbf{\tdataset}: a new \textit{\textbf{V}irtual \textbf{R}eality dataset for
\textbf{Biom}etric Applications} acquired from the \tdevice. One subset of this
dataset comprises 900 short videos, each lasting nearly 10 seconds, collected
from 25 individuals under various conditions, including steady gaze, moving
gaze, and half-closed eye settings. To ensure evaluation and development of
various use-cases in realistic scenarios, half of these videos were recorded
with subjects wearing glasses, while the remaining half were recorded without
glasses. These videos present potential for a wide range of applications, such
as biometric recognition and semantic segmentation. It should be noted that
conventional biometric recognition pipelines comprise of sequence of tasks,
including region of interest (RoI) detection, segmentation, and orientation. The
presented dataset may be harnessed to enhance each of these tasks separately.
We have also captured recordings using different types of 3D masks, combined
with synthetic eyes such as 2D print-outs and 3D fake eyes. These 1104
recordings, also called as presentation attacks (PAs), comprise other subset of
the \tdataset dataset meant for addressing concerns related to PAD and
anti-spoofing measures. The overall \tdataset dataset consists of 2004
recordings of periocular (left and right) regions captured in near-infrared
(NIR) spectrum for 10$s$ at 72 FPS.

The present release of the \tdataset consists of videos and subject-level
(identity) labels. It also provides details of the scenarios and PA instruments
used. We do not provide pixel-level annotations for semantic details.  The
contributions of our work can be summarized as below:
\begin{itemize}[leftmargin=*, itemsep=0pt, topsep=0pt]
\item We have created and publicly released the \tdataset dataset\footnote{The
\tdataset dataset can be downloaded from
\url{https://www.idiap.ch/dataset/vrbiom}.},
a collection of more than 2000 periocular videos acquired from a VR device. To
the best of our knowledge, this is the first publicly available dataset
featuring realistic, non-frontal views of the periocular region for a variety of
biometric applications.
\item The \tdataset dataset offers a range of realistic and challenging
variations for the benchmarking and development of biometric applications. It
includes recordings under three gaze conditions: steady gaze, moving gaze, and
partially closed eyes. Additionally, it provides an equal split of recordings
with and without glasses, facilitating the analysis of the impact of eyewear on
iris/ periocular recognition.
\item As a part of \tdataset, we have also released more than 1100 PA videos
constructed from 92 attack instruments. These attacks, incorporating different
combinations of PAIs such as prints, fake 3D eyeballs, and various masks,
provide a valuable resource for advancing PAD research in a VR setup.
\end{itemize}

The organization of this paper is structured as follows:
Section~\ref{sec:related_work} provides a brief review of iris/periocular
datasets and HMD-based datasets. Section~\ref{sec:dataset} explains details of
the \tdataset dataset, including data collection methods and associated
challenges. In Section \ref{sec:use_cases}, we discuss potential use cases of
the dataset. Finally, we summarize the work with Section~\ref{sec:conc}.


%% file: sections/fig_usecase_collection.tex
\begin{figure}[h]
\centering
\includegraphics[width=0.6\textwidth]{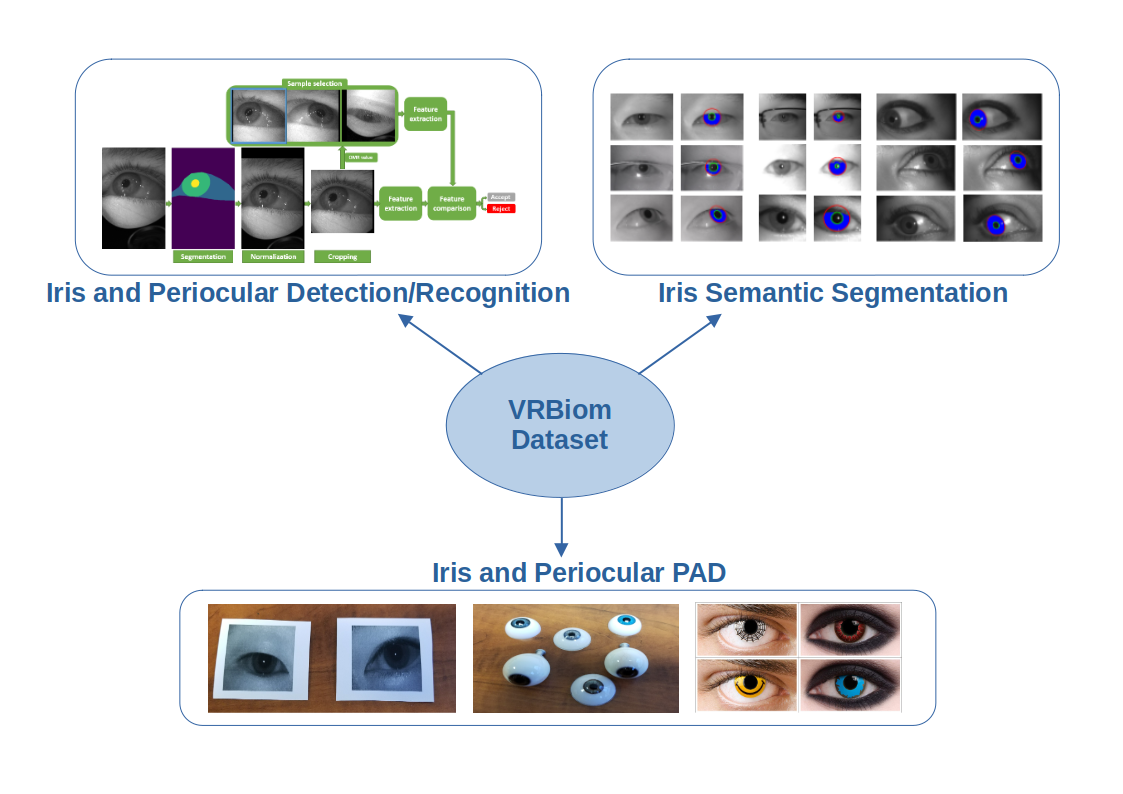}
\caption{Examples of biometric use-cases offered by the HMD data, in particular by
the \tdataset dataset (The example images are obtained from~\cite{9107939,
wang2021nir, hoffman2019iris}).}
\label{fig:use_cases}
\end{figure}

%% file: sections/related_work.tex
In this section, we briefly present some commonly used datasets for
iris/periocular biometrics, followed by review of recent datasets collected
using HMD devices.

\subsection{Datasets for Iris/ Periocular Biometrics}

One of the most prominent and commonly used iris datasets is the CASIA-Iris
series\footnote{\url{http://www.cbsr.ia.ac.cn/english/Databases.asp}}, collected
by the Chinese Academy of Sciences, which consists of four main versions denoted
with suffixes V1–V4~\cite{casia_iris}. The first version, V1, released in 2002,
comprised 756 images taken from 108 subjects at a resolution of $320 \times
280$. In 2004, CASIA-IrisV2 was released, containing 2,400 images from 120
individuals at VGA resolution ($640 \times 480$). The third version included
three different subsets: Interval, Lamp, and Twins. The Casia-Interval subset
was captured from 249 subjects in an indoor environment, while the Lamp subset
provided 16$k$ images at VGA resolution. The Twins subset contained recordings
of 200 twins, primarily children, in an outdoor
environment~\cite{zanlorensi2022ocular}. The latest version, CASIA-IrisV4, is
the most comprehensive, including 54,601 images from 2,800 subjects distributed
across six subsets namely CASIA-Iris-Interval, Lamp, Twins, Distance, Thousand,
and Syn~\cite{casia_iris, omelina2021survey}. 

The Iris Liveness Detection Competition (LivDet-Iris)~\cite{yambay2017livdet,
tinsley2023iris} dataset combines multiple datasets designed for liveness
detection in the context of iris PAD. The 2017 competition utilized four
different datasets created by Clarkson, Warsaw, Notre Dame (ND), Indraprastha
Institute of Information Technology Delhi (IIIT-Delhi), and West Virginia
University (WVU)~\cite{yambay2017livdet}. The Clarkson and Warsaw datasets
mainly focus on print attacks, consisting of 8,095 and 12,013 images,
respectively. The ND CLD 2015 and IIITD-WVU datasets comprise more than 7$k$
images each, acquired at VGA resolution. The recent edition of the competition,
held in conjunction with IJCB 2023, aimed at creating more challenging
attacks~\cite{tinsley2023iris}. This dataset included attack instruments such as
print, contact lenses, electronic displays, fake/prosthetic eyes, and synthetic
iris of varying quality.

Researchers at the University of Beira have released a series of UBIRIS datasets
of iris images. The UBIRIS v1 dataset provides 1,877 images acquired under less
constrained imaging conditions~\cite{proencca2005ubiris}, whereas UBIRIS v2
offers more than 11$k$ images captured at a distance~\cite{proencca2009ubiris}.
Another version, UBIPr, derived from UBIRIS v2, provides wider regions suitable
for periocular recognition. They have also released pixel-level masks of
semantic elements such as the iris and sclera, making it a useful resource for
segmentation applications.

Additionally, the Multimedia University (MMU) released two iris image datasets
in 2010 that are publicly available. The first version, MMU-v1, consists of 450
images obtained from 45 subjects, while the subsequently released MMU-v2
provides 995 images from 100 subjects~\cite{teo2010robust, nguyen2024deep}.

Table~\ref{tab:iris_datasets} summarizes the iris and periocular biometric
datasets, providing details related to samples and attacks where applicable.

\input{sections/table_iris_datasets}


\subsection{HMD Datasets for Iris/ Periocular Biometrics}

Due to the inherent challenges in collecting data using HMD devices and the
limitations of hardware quality, there have been fewer attempts to gather such
datasets. However, with recent advancements in hardware technology and the
increasing number of companies releasing HMD headsets, these types of datasets
are gaining traction in both research and end-user applications.

Released in 2012, the Point of Gaze (PoG) dataset is one of the HMD datasets
created for gaze detection and head pose studies~\cite{mcmurrough2013dataset}.
It comprises data from 20 subjects recorded at a resolution of $768 \times 480$.
The Labelled Pupils in the Wild (LPW) dataset contains 66 eye-region videos at
VGA resolution captured at 95 FPS~\cite{tonsen2016labelled}. As its name
suggests, the LPW dataset features several variations in illumination, eyewear
(glasses and contact lenses), and makeup, along with gaze directions.
Kim~\textit{et al.} introduced the NVGaze dataset, which was collected from 35
subjects~\cite{kim2019nvgaze}. This dataset focuses on gaze estimation and
includes approximately 2.5 million \tbf images at VGA resolution.

The Open Eye Dataset (OpenEDS) has two editions: OpenEDS 2019 and 2020, both
released by Facebook Reality Labs, which have significantly advanced the field
of HMD-based eye-tracking research. The OpenEDS 2019
dataset~\cite{garbin2019openeds} comprises videos and images from 152 subjects
and is partially annotated for semantic segmentation. The 2020
version~\cite{palmero2020openeds2020} includes two sub-datasets (the Gaze
Prediction and Eye Segmentation Datasets) from 80 subjects, comprising
approximately 580$k$ images with a resolution of $640 \times 400$ pixels,
captured at 100 FPS.

The creators of OpenEDS also organized a competition (challenge) to improve
performance in tracking and segmentation tasks. While OpenEDS is similar to our
presented work, it differs in two main aspects: first, OpenEDS data is acquired
from eye-facing cameras, yielding frontal eye views, whereas our dataset uses a
consumer-level HMD capturing eyes and periocular regions from non-frontal
angles. Second, OpenEDS includes recordings only from real (\tbf) subjects,
while our dataset incorporates a variety of presentation attacks (PAs), making
it valuable for PAD research as well~\cite{garbin2019openeds,
palmero2020openeds2020}.

Our dataset, \tdataset, consists of 900 short videos from 25 participants and
more than 1100 videos across six categories of attack instruments. Each video,
with a spatial resolution of $400 \times 400$, is approximately 10 seconds long
and captured at 72 FPS. Table~\ref{tab:hmd_datasets} provides a brief comparison
of the publicly available HMD datasets for research.

\input{sections/table_hmd_datasets}


%% file: sections/table_iris_datasets.tex
\begin{table}[h]
\renewcommand{\arraystretch}{1.4}
\centering
\resizebox{\linewidth}{!}{%
\begin{tabular}{l| l | l | l | l | p{3cm} | p{1.5cm} | l } 
\hline
\multicolumn{2}{c|}{\textbf{Datasets and Subsets}} &   \textbf{\#Subjects} &
\textbf{\#Images} &  \textbf{Resolution} &  \textbf{Types of Attacks} &
\textbf{Collection Year} &  \textbf{Institution(s)} \\ \hline
\multirow{4}{*}{CASIA-Iris series~\cite{casia_iris}} 
& CASIA-IrisV1 &    108*    & 756    & $320 \times 280$ & No     & 2002 &  \multirow{4}{*}{Chinese Academy of Sciences} \\
& CASIA-IrisV2 &    120**   & 2,400  & $640 \times 480$ & No     & 2004 &         \\
& CASIA-IrisV3 &    700     & 22,034 & Various          & No     & 2005 &         \\
& CASIA-IrisV4 &    2,800   & 54,601 & Various          & No     & 2010 &         \\
\hline
\multirow{4}{*}{LivDetIris 2017~\cite{yambay2017livdet}} 
& ND CLD 2015 & -   & 7,300     & $640 \times 480$ & contact lens (live)                  & 2015 & University of Notre Dame    \\
& IIITD-WVU   & -   & 7,459     & $640 \times 480$ & Print                                & 2017 & IIIT Delhi   \\
& Clarkson    & 50  & 8,095     & $640 \times 480$ & Print and contact lens (live)        & 2017 & Clarkson University \\
& Warsaw      & 457 & 12,013    & $640 \times 480$ & Print and contact lens (print, live) & 2017 & Warsaw University \\
\hline
\multirow{2}{*}{UBIRIS~\cite{proencca2005ubiris,proencca2009ubiris}} 
& UBIRIS-V1 & 241 &  1,877 &  $400 \times 300$ &  No & 2004 & \multirow{2}{*}{University of Beira} \\
& UBIRIS-V2 & 261 & 11,102 &  $800 \times 600$ &  No & 2009 &  \\
\hline
\multirow{2}{*}{MMU~\cite{teo2010robust, nguyen2024deep}}
& MMU-V1       & 45  & 450    & -    & No   & 2010 & \multirow{2}{*}{Multimedia University} \\
& MMU-V2       & 100 & 995    & -    & No   & 2010 &          \\
\hline
LivDet-Iris 2023~\cite{tinsley2023iris} 
& - & - & 13,332*** & Various & Print, contact lens, fake/ prosthetic eyes, and synthetic iris & 2023 & Multiple Institutions\\
\hline
\end{tabular}
}
\caption{Overview of iris and periocular biometric datasets. (*number of classes, **number of eyes, ***number of test data).}
\label{tab:iris_datasets}
\end{table}

%% file: sections/table_hmd_datasets.tex
\begin{table}[h]
\renewcommand{\arraystretch}{1.4}
\centering
\resizebox{\linewidth}{!}{%
\begin{tabular}{l| p{1.5cm}| l |l |l |l | p{2cm}| l| p{1.5cm}}
\hline
\textbf{Datasets} & \textbf{Collection Year} & \textbf{\#Subjects} &
\textbf{\#Images} & \textbf{Resolution} & \textbf{FPS} (Hz) &
\textbf{Presentation Attacks}  & \textbf{Profile} & \textbf{Synthetic
        Eye Data} \\ \hline
Point of Gaze (PoG)~\cite{mcmurrough2013dataset}          & 2012 & 20  & --            & $768 \times 480$ & 30  & \xmark   & Frontal     & \xmark  \\
LPW~\cite{tonsen2016labelled}                             & 2016 & 22  & 130,856       & $640 \times 480$ & 95  & \xmark   & Frontal     & \xmark  \\
NVGaze~\cite{kim2019nvgaze}                               & 2019 & 35  & 2,500,000     & $640 \times 480$ & 30  & \xmark   & Frontal     & \cmark \\
OpenEDS2019~\cite{garbin2019openeds}                      & 2019 & 152 & 356,649       & $640 \times 400$ & 200 & \xmark   & Frontal     & \xmark  \\
OpenEDS2020~\cite{palmero2020openeds2020}                 & 2020 & 80  & 579,900       & $640 \times 400$ & 100 & \xmark   & Frontal     & \xmark  \\\hline
\textbf{\tdataset (this work)}                            & 2024 & 25  & 1,262,520     & $400 \times 400$ & 72  & \cmark   & Non-Frontal & \cmark \\
\hline
\end{tabular}
}
\caption{Summary of the iris/ periocular datasets acquired by the HMDs.}
\label{tab:hmd_datasets}
\end{table}
    

%% file: sections/dataset.tex
%
In this section, we describe the process of acquiring both \tbf and PA data,
along with relevant statistics. We also discuss some of the challenges
encountered in the creation of the \tdataset dataset. 
 
\subsection{Collection Setup}
A total of 25 subjects, aged between 18 and 50 and representing a diverse range
of skin tones and eye colors, participated in the data collection process. Each
participant was briefed on the project objectives and provided their consent
through a signed consent form. The data acquisition for each participant was
conducted in a single session lasting approximately 20 minutes. We divided the
recordings into two sub-sessions: the first without the subject wearing glasses
and the second with glasses. For each sub-session, we captured three recordings
involving different gaze variations: steady gaze, moving gaze, and partially
closed eyes. Each video had a duration of approximately 10 seconds, recorded at
a frame rate of 72 frames per second (FPS). The videos have a spatial resolution
of $400 \times 400$ pixels and were captured in a single-channel in the NIR
spectrum. It was often observed that for nearly first second of the recording,
the recordings were over-exposed to the NIR illumination. Therefore, we
recommend discarding the first 70--80 frames from processing.

\begin{figure}[h]
\begin{subfigure}{0.3\columnwidth}
\includegraphics[width=\textwidth]{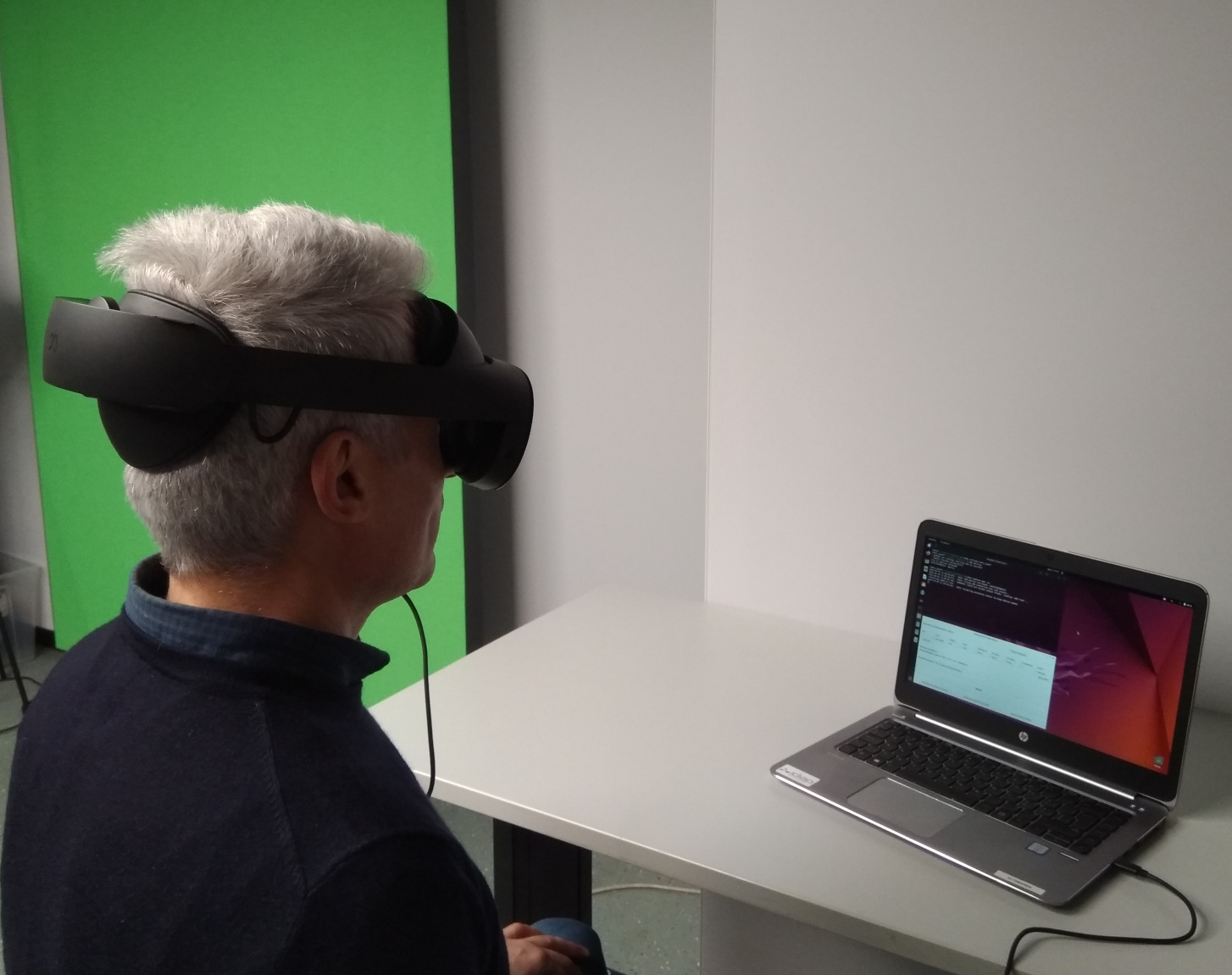}
\caption{}
\label{fig:bf_recording}
\end{subfigure}
\quad
\begin{subfigure}{0.3\columnwidth}
\includegraphics[width=\textwidth]{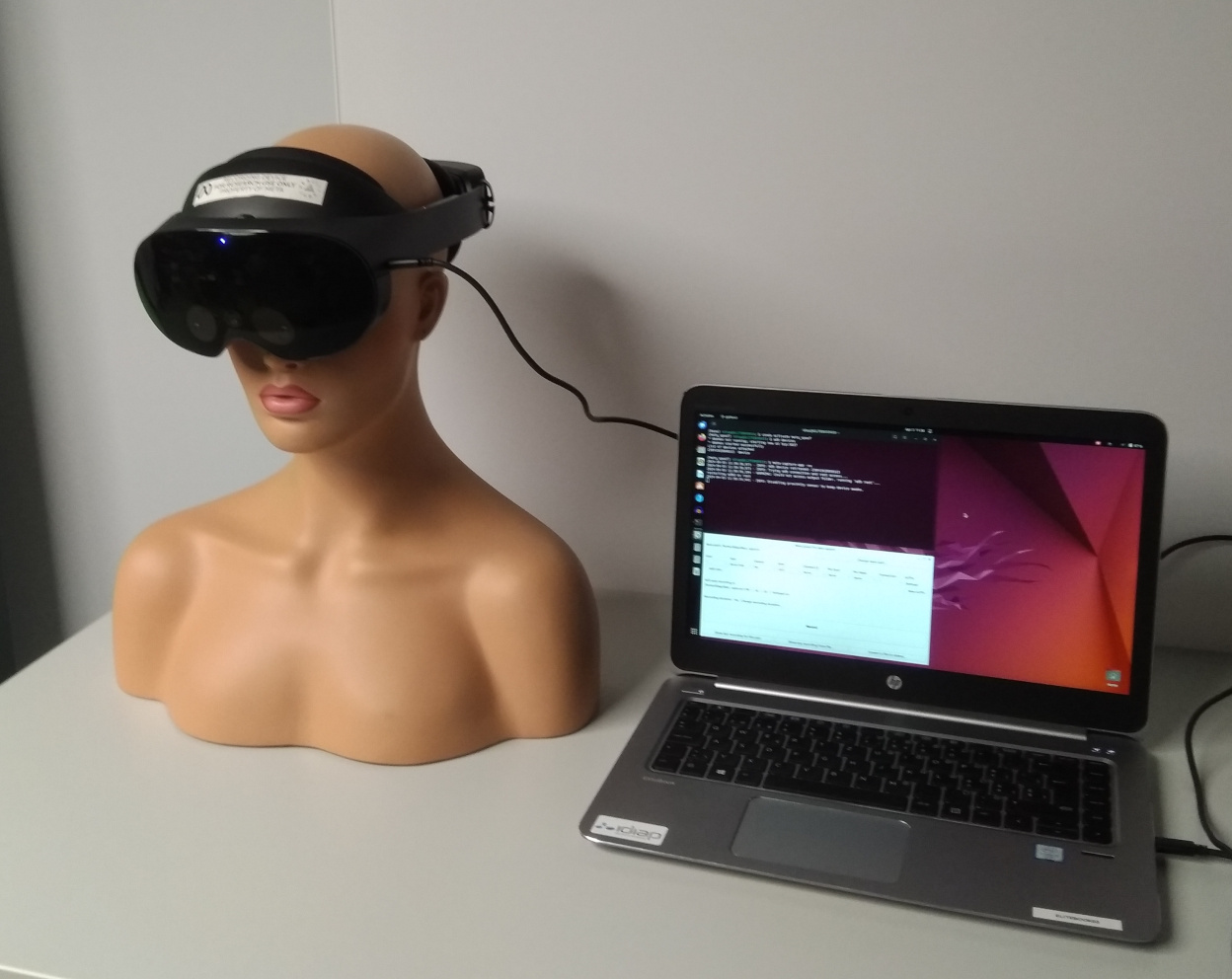}
\caption{}
\label{fig:pa_recording}
\end{subfigure}
\quad
\begin{subfigure}{0.35\columnwidth}
\includegraphics[width=\textwidth]{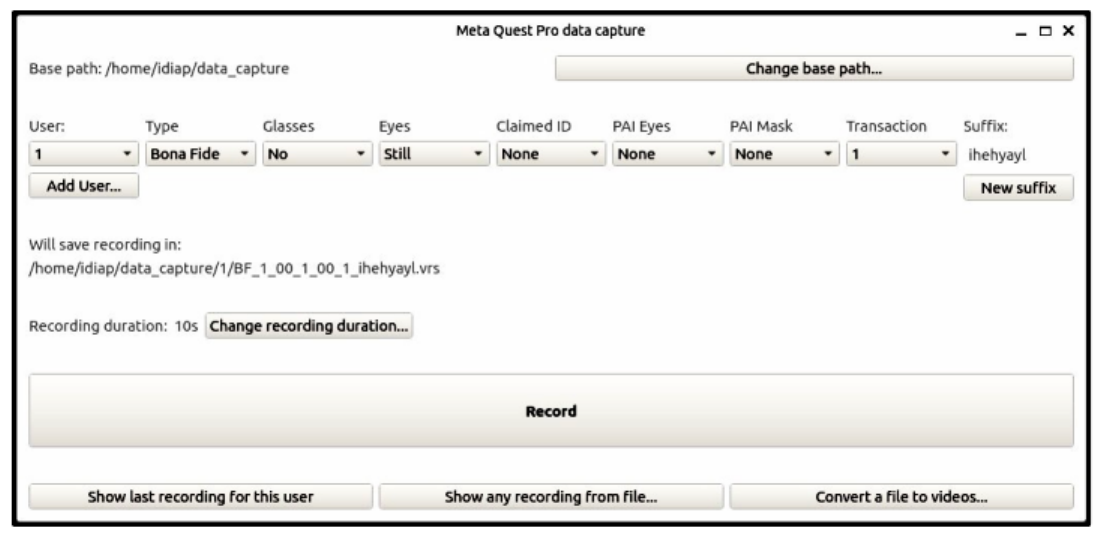}
\caption{}
\label{fig:capture_app}
\end{subfigure}
\caption{Setup for the dataset collections: (a) for \tbf recordings, the
subjects wore the HMD devices, (b) for PA recordings, the HMD device was
carefully placed on the temple region of the attack instrument
(mannequin, in this example), and (c) the Meta Quest Pro Data Capture
App used for data collection.}
\end{figure}

The iris/periocular videos in the \tdataset dataset were captured by the inward
facing cameras located in the \tdevice headset. For \tbf recordings, the
subjects wore the HMD device, whereas for recording of PAs, we systematically
positioned the \tdevice on the attack instrument (as described later in this
section). The \tdevice was connected to a workstation via USB-C cable which
facilitated both charging and the transfer of recording commands and data.
Figs.~\ref{fig:bf_recording} and~\ref{fig:pa_recording} depict the recording
setups for \tbf and PAs, respectively.

To streamline the data collection, we developed the Meta Quest Pro Data Capture
App (app) (Refer to Fig.~\ref{fig:capture_app} for GUI). This app enabled us to
set file names, manage recordings, and transfer files from the \tdevice to our
workstation efficiently. Additionally, the app provided tools to convert the
video recordings from the proprietary \textit{vrs} format to the conventional
\textit{avi} format, which is compatible with most common video players and
softwares. Through the capture app, we also ensured a specific naming convention
for each recoded video. The naming convention consisted of the following fields:
Type (\tbf or PA), subject ID (or PA ID), Eye (left or right), glasses (without
or with), type of gaze (Steady, Moving, or Partially closed), claimed ID (for
PAs), details of PAI (for PAs), recording ID, and a random suffix. These
fields, with their mapped values, were used to create the filename for each
video. More details on this can be found in the actual dataset.


\subsection{\tbf Recordings}

Before starting the recording, we ensured that the HMD (\tdevice) is securely
and comfortably fitted around the subject's head, providing them with an
immersive virtual experience. Considering the diverse array of use-cases, we
captured videos under three different gaze variations:

\begin{itemize}[leftmargin=10pt, itemsep=0pt, topsep=0pt]
\item \textbf{Steady Gaze:} The subject maintains a nearly fixed gaze position
by fixating their eyes on a specific (virtual) object.
\item \textbf{Moving Gaze:} The subject's gaze moves freely across the scene.
\item \textbf{Partially Closed Eyes:} The subject keeps their eyes partially
closed without focusing on any particular gaze.
\end{itemize}

\input{sections/fig_bf_samples}%

These variations were recorded under two eye-wear conditions: with glasses and
without glasses. If the subjects did not have their own medical glasses, they
were provided with a pair of fake glasses. Each gaze variation and glasses
condition was repeated three times to ensure a robust dataset.

Fig.~\ref{fig:bf_samples} provides samples of each of the aforementioned
variations. The HMD is equipped with cameras that simultaneously record videos
of both eyes. Therefore, for each recording, two videos are captured
concurrently- one for the left eye and one for the right eye. In total, we
collected 900 \tbf videos from 25 subjects, which can be summarized as follows:

\begin{table}[]
\renewcommand{\arraystretch}{2}
\centering
\begin{tabular}{l| l | l | l| l | l} \hline
\textbf{\#Subjects} & \textbf{\#gaze} & \textbf{\#glass} & \textbf{\#repetitions} & \textbf{\#eyes} & \textbf{\#total} \\ \hline
25         & 3      & 2       & 3             & 2      & 900 \\ \hline
\end{tabular}
\caption{Summary of \tbf recordings in the \tdataset dataset.}
\label{tab:bf_summary}
\end{table}

Thus, the structure and coverage of the \tdataset ensure that each subject's
iris/periocular region is recorded across various realistic scenarios, providing
a valuable resource for researchers to analyse and develop biometric applications.

\subsection{PA Recordings}
Most of the PAs were constructed by combining two categories of attack
instruments: those targeting the eyes and those targeting the periocular region.
For the eyes, we used a variety of instruments including fake 3D eyes
(eyeballs), printouts from synthetic and real identities, and plastic-made
synthetic eyes. For the periocular region, we employed mannequins and 3D masks
made of different materials. These masks also served as a fake head where the
HMD could be securely placed for recording.

\input{sections/fig_pa_samples}%

Using these combinations, we created six categories of PAIs, each with a
variable number of attack instruments. Samples of these categories are provided
in Fig.~\ref{fig:pa_samples}, with a brief description as follows:

\begin{itemize}[leftmargin=10pt, itemsep=0pt, topsep=0pt]
\item \textbf{Mannequins:} We used a collection of seven mannequins made of
plastic, each featuring its own eyes. These mannequins are generic (fake heads)
as they do not represent any specific \tbf identity. Unlike the requirements of
additional components for other PAIs to provide a stable platform or extra
equipment to stabilize HMDs, these mannequins are particularly advantageous due
to their integrated and stable platform for placing HMDs.

\item \textbf{Custom rigid masks (type-I):} This category includes ten custom
rigid masks, comprising five men and five women. The term custom implies that
the masks were modeled after real individuals. Manufactured by REAL-f Co.\ Ltd.\
(Japan), these masks are made from a mixture of sandstone powder and resin. They
represent real persons (though not part of the \tbf subjects in this work) and
include eyes made of similar material with synthetic eyelashes attached.

\item \textbf{Custom rigid masks (type-II):} We used another collection of 14
custom rigid masks to construct the PAs for the \tdataset. These masks differ
from the previous category in two ways: first, they are made of amorphous powder
compacted with resin material by Dig:ED (Germany); second, they have empty
spaces at the eye locations, where we inserted fake 3D glass eyeballs to
construct an attack.

\item \textbf{Generic flexible masks:} This category includes twenty flexible
masks made of silicone. These full-head masks do not represent any specific
identity, hence termed generic. The masks have empty holes for eyes, where we
inserted printouts of periocular regions from synthetic identities. Using
synthetic identities alleviates privacy concerns typically associated with
creating biometric PAs.

\item \textbf{Custom flexible masks:} These silicone masks represent real
individuals. Similar to the previous attack categories, these masks have holes
at the eye locations, where we inserted fake 3D eyes to create the attacks. We
have 16 PAIs in this category.

\item \textbf{Vulnerability Attacks:} For each subject in the \tbf collection,
we created a print attack. For both eye-wear conditions (without and with
glasses), we manually selected an appropriate frame from each subject's \tbf
videos and printed it at true-scale on a laser printer. The printouts were cut
into periocular crops and placed on a mannequin to resemble a realistic
appearance when viewed by the tracking cameras of the HMD. The so-obtained 25
print PAs can be used to assess the vulnerability of the corresponding biometric
recognition system. These attacks simulate scenarios where an attacker gains
access to (unencrypted) data of authorized individuals and constructs simple
attacks using printouts.

\end{itemize}

Table~\ref{tab:dataset_details} summarizes the details of the \tdataset dataset
comprising 900 \tbf and 1104 PA videos. It also provides the naming conventions
used to indicate the type of PAI.

\begin{table*}[h]
\renewcommand{\arraystretch}{1.4}
\centering
\begin{tabular}{p{1.7cm} | p{1.5cm} | p{4cm} | p{1.7cm} | p{1.7cm} | p{4cm}} \hline
\textbf{Type}  & \textbf{PA Series} & \textbf{Subtype}       & \textbf{\# Identities} & \textbf{\# Videos} & \textbf{Attack Types}          \\ \hline
\tbf  & --     & [steady gaze, moving gaze, partially closed] $\times$ [glass, no glass] & 25   & 900  & --   \\ \hline
\multirow{6}{2cm}{Presentation Attacks}
    & 2  & Mannequins                    & 7     & 84    & Own eyes (same material)       \\
    & 3  & Custom rigid mask (type I)    & 10    & 120   & Own eyes (same material)       \\
    & 4  & Custom rigid mask (type II)   & 14    & 168   & Fake 3D eyeballs               \\
    & 5  & Generic flexible masks        & 20    & 240   & Print attacks (synthetic data) \\
    & 6  & Custom silicone masks         & 16    & 192   & Fake 3D eyeballs               \\
    & 7  & Print attacks                 & 25    & 300   & Print attacks (real data)      \\ \hline
\end{tabular}
\caption{Details of \tbf and different types of PAs from the \tdataset dataset. Each video
was recorded at 72 FPS for approximately 10$s$.}
\label{tab:dataset_details}
\end{table*}

\subsection{Challenges}

Here we briefly discuss some of the challenges encountered during data
collection, and the corresponding solutions employed. 
 
\begin{itemize}[leftmargin=10pt, itemsep=0pt, topsep=0pt]
\item Eyelashes: During the initial data capture experiments, we observed that
the subjects' eyelashes appeared as a prominent feature, especially when
capturing the periocular region. The absence of eyelashes in most PAIs, where
prints or 3D eyeballs were used to construct the attack, can be considered as an
easy to distinguish yet unrealistic feature. To create a more realistic
scenario, we decided to use false eyelashes from standard makeup kits.

\item PA recordings: Recording the PAs was challenging due to the lack of
real-time feedback from the device or capture app. To achieve the correct
positioning and angle for various PAIs, it often required multiple attempts and
recording trials for data collectors to ensure the desired quality of the
captured data.
 
\item NIR Camera's Over-exposure: The proximity between the attack instrument
and the NIR illuminator/receiver often resulted in over-exposed recordings.
Given that VR headsets are designed to fit closely on the forehead, we had a
limited room for adjustment. However, we attempted minor adjustments to the
HMD's orientation and the placement of the PAI to mitigate exposure issues to a
considerable extent.
 
\item Synthetic Eyes: One challenge we faced was the variability in the identity
of synthetic eyes. To ensure accurate biometric measurements, we needed to
carefully select synthetic eyes that closely resemble human eyes. 
 
\item Print Attack: The print-out attacks designed for vulnerability assessment
are inherently 2-dimensional, whereas the periocular region's appearance is 3D.
To simulate realistic appearance, we employed small paper balls to roll the
printed attacks, thereby providing a structural view akin to 3D representation.
 
\end{itemize}

%
%

%% file: sections/fig_bf_samples.tex
\begin{figure}[h]
\centering
\begin{subfigure}{0.3\columnwidth}
\includegraphics[width=\textwidth]{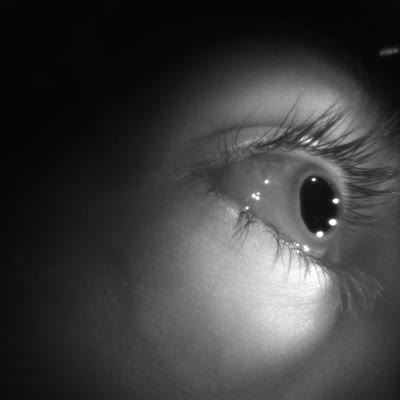}
\end{subfigure}
\,
\begin{subfigure}{0.3\columnwidth}
\includegraphics[width=\textwidth]{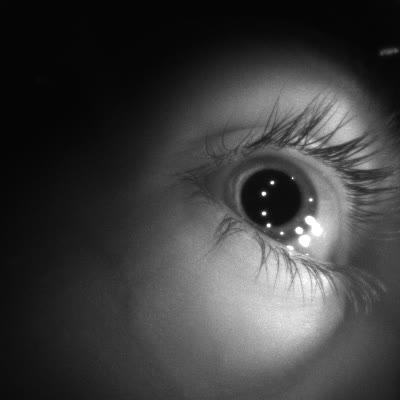}
\end{subfigure}
\,
\begin{subfigure}{0.3\columnwidth}
\includegraphics[width=\textwidth]{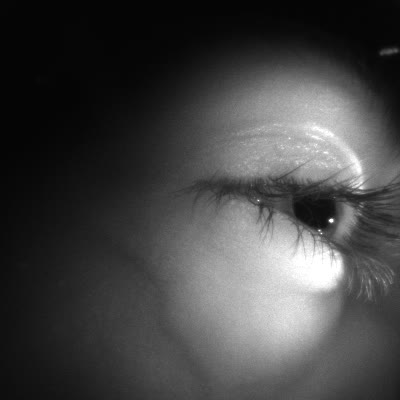}
\end{subfigure}
\\
\medskip
\begin{subfigure}{0.3\columnwidth}
\includegraphics[width=\textwidth]{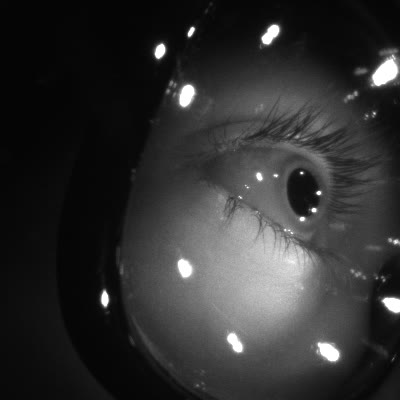}
\end{subfigure}
\,
\begin{subfigure}{0.3\columnwidth}
\includegraphics[width=\textwidth]{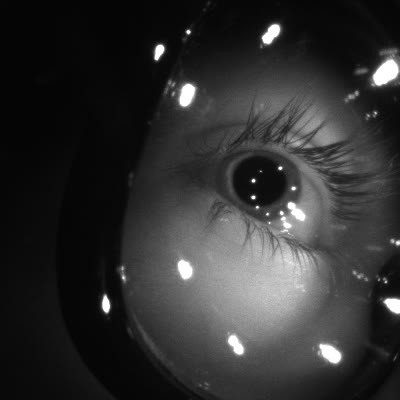}
\end{subfigure}
\,
\begin{subfigure}{0.3\columnwidth}
\includegraphics[width=\textwidth]{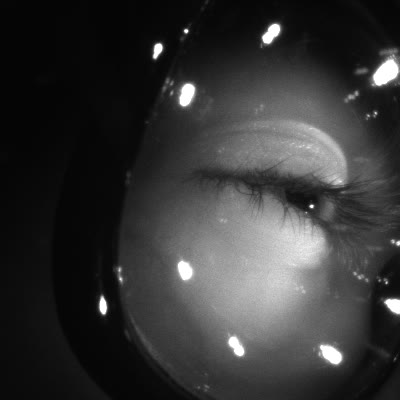}
\end{subfigure}
\caption{Samples of \tbf recordings from \tdataset dataset. Each row presents a sample
of steady gaze, moving gaze, and partially closed eyes (from left to right). Top
and bottom rows refer to recordings without and with glasses, respectively.}
\label{fig:bf_samples}
\end{figure}

%% file: sections/fig_pa_samples.tex
\begin{figure}[h]
\centering
\begin{subfigure}{0.15\columnwidth}
\includegraphics[width=\textwidth]{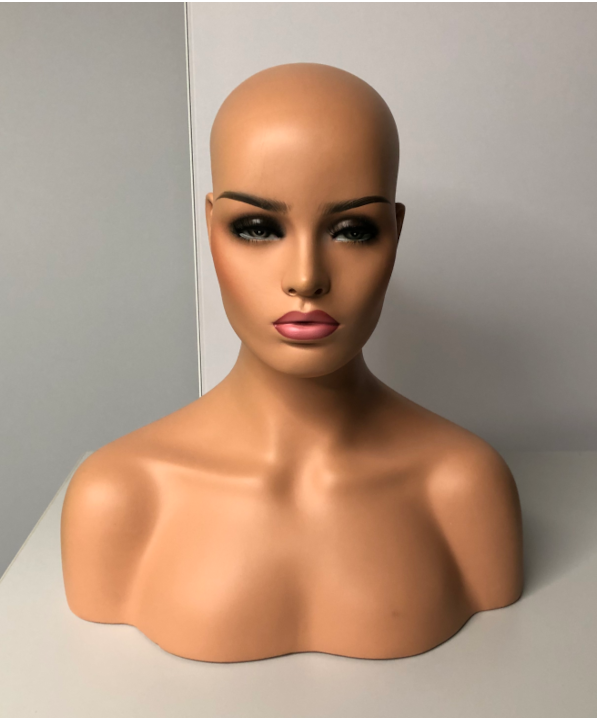}
\end{subfigure}
\,
\begin{subfigure}{0.15\columnwidth}
\includegraphics[width=\textwidth]{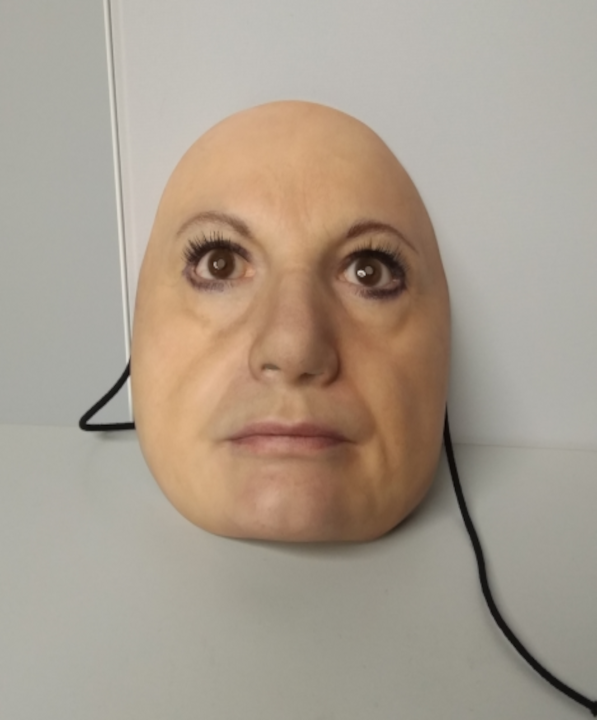}
\end{subfigure}
\,
\begin{subfigure}{0.15\columnwidth}
\includegraphics[width=\textwidth]{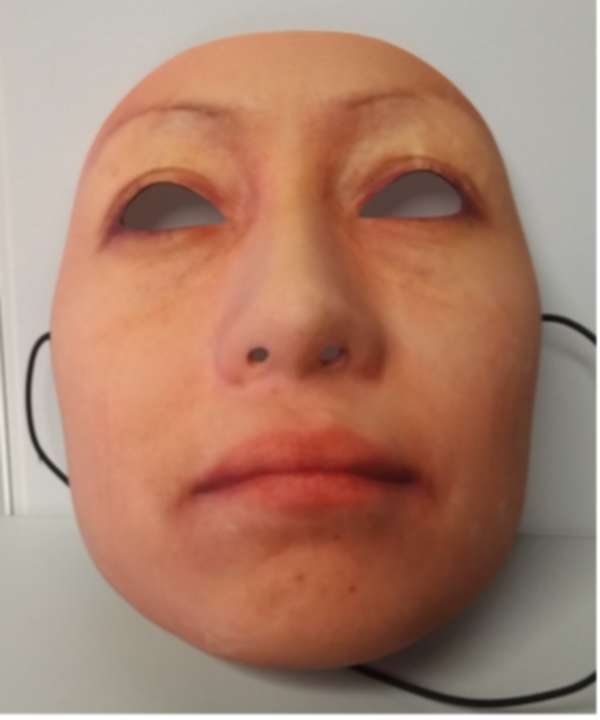}
\end{subfigure}
\,
\begin{subfigure}{0.15\columnwidth}
\includegraphics[width=\textwidth]{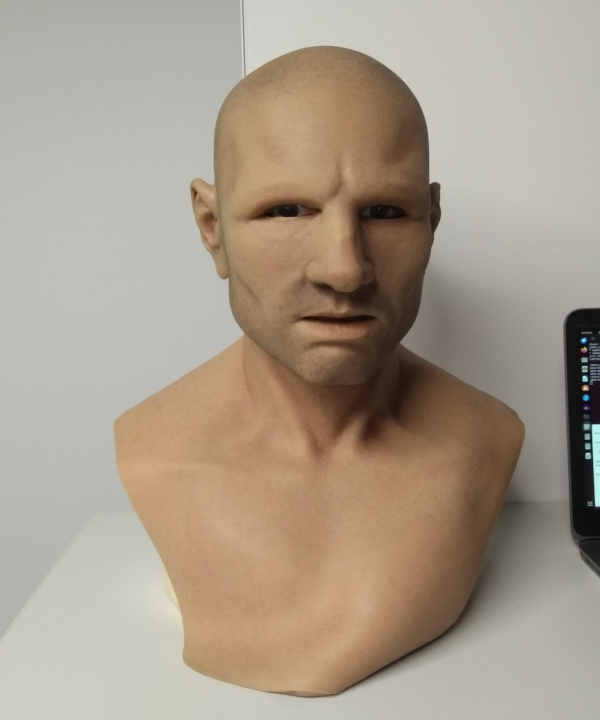}
\end{subfigure}
\,
\begin{subfigure}{0.15\columnwidth}
\includegraphics[width=\textwidth]{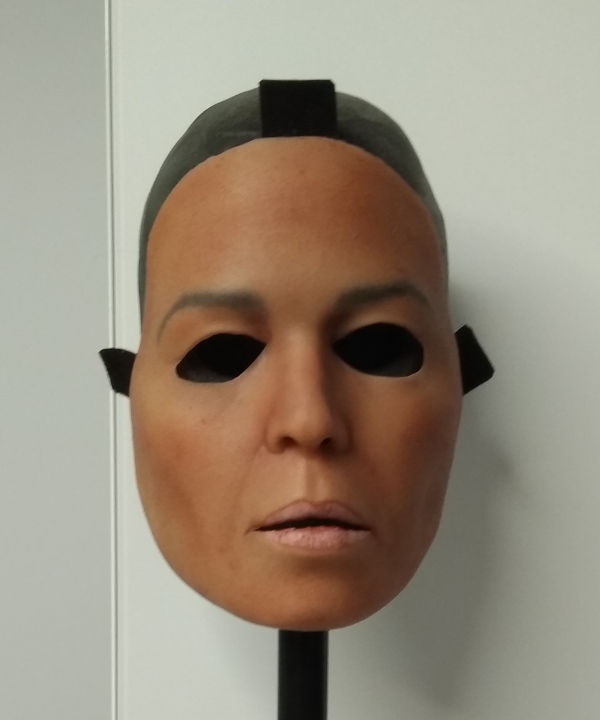}
\end{subfigure}
\,
\begin{subfigure}{0.15\columnwidth}
\includegraphics[width=\textwidth]{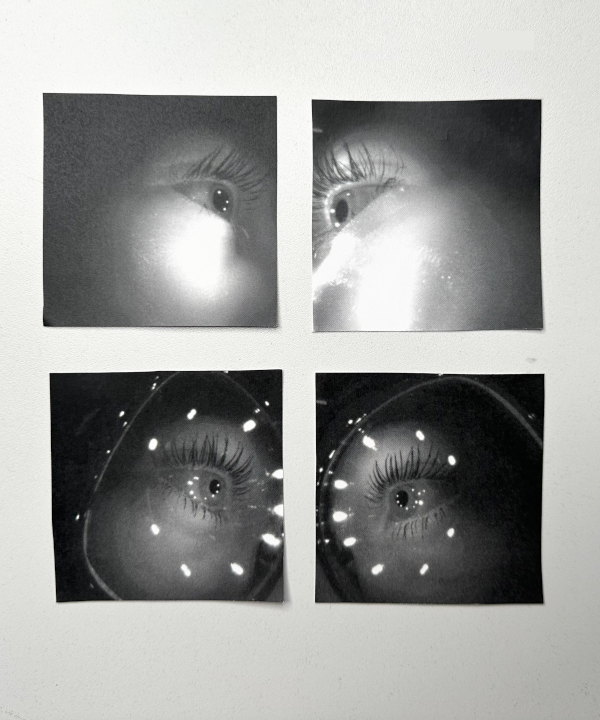}
\end{subfigure}
\\
\bigskip
\begin{subfigure}{0.15\columnwidth}
\includegraphics[width=\textwidth]{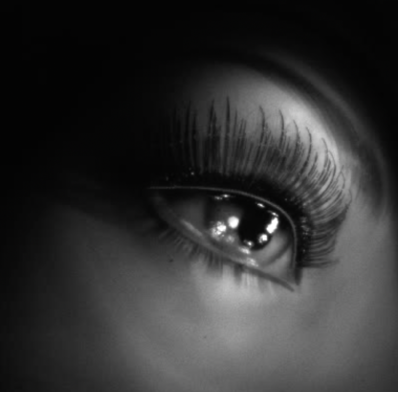}
\end{subfigure}
\,
\begin{subfigure}{0.15\columnwidth}
\includegraphics[width=\textwidth]{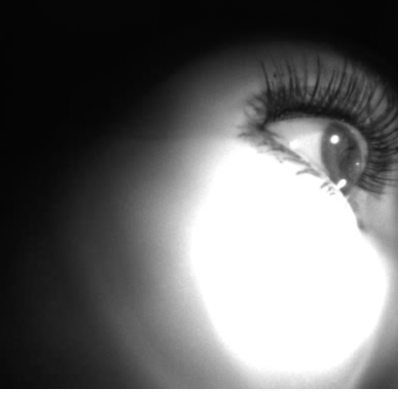}
\end{subfigure}
\,
\begin{subfigure}{0.15\columnwidth}
\includegraphics[width=\textwidth]{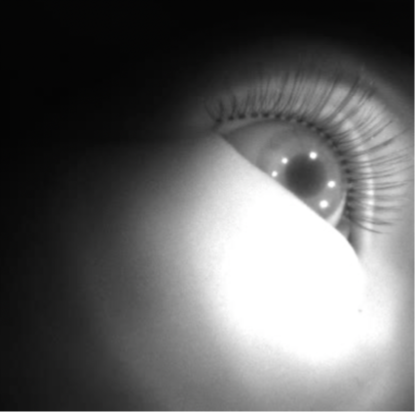}
\end{subfigure}
\,
\begin{subfigure}{0.15\columnwidth}
\includegraphics[width=\textwidth]{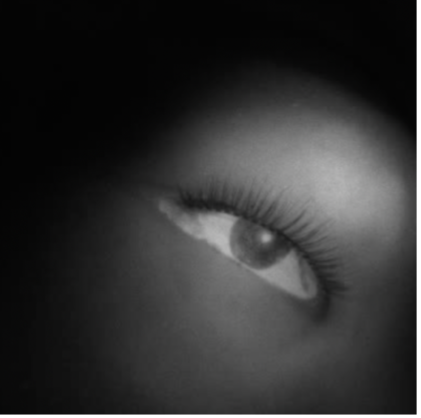}
\end{subfigure}
\,
\begin{subfigure}{0.15\columnwidth}
\includegraphics[width=\textwidth]{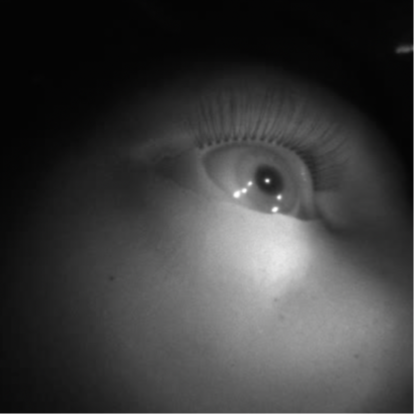}
\end{subfigure}
\,
\begin{subfigure}{0.15\columnwidth}
\includegraphics[width=\textwidth]{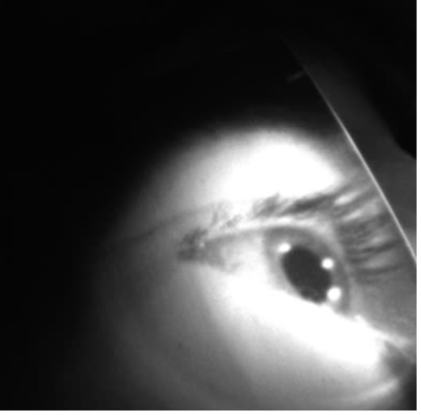}
\end{subfigure}
\\
\medskip
\begin{subfigure}{0.15\columnwidth}
\includegraphics[width=\textwidth]{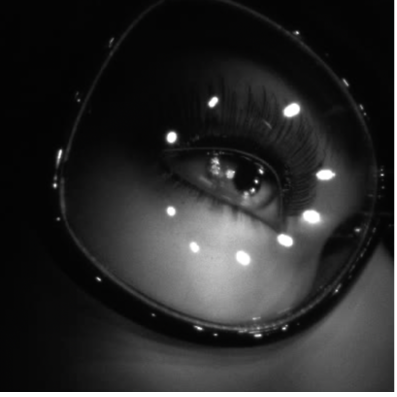}
\end{subfigure}
\,
\begin{subfigure}{0.15\columnwidth}
\includegraphics[width=\textwidth]{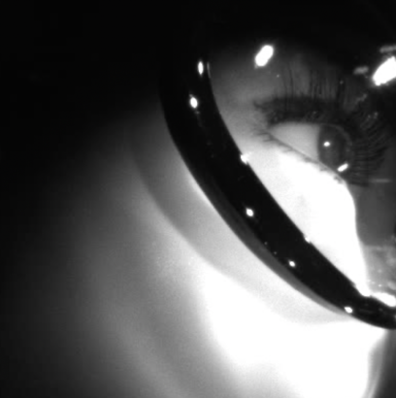}
\end{subfigure}
\,
\begin{subfigure}{0.15\columnwidth}
\includegraphics[width=\textwidth]{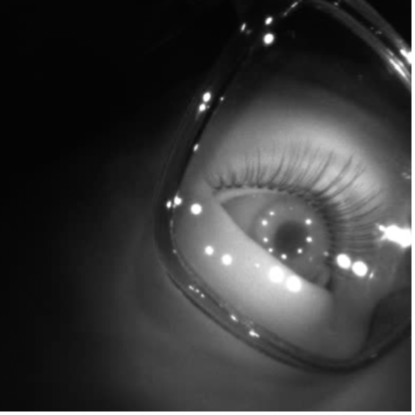}
\end{subfigure}
\,
\begin{subfigure}{0.15\columnwidth}
\includegraphics[width=\textwidth]{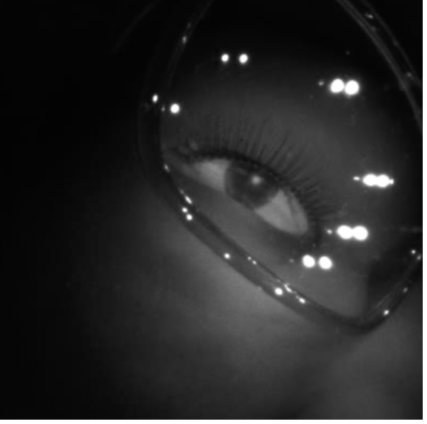}
\end{subfigure}
\,
\begin{subfigure}{0.15\columnwidth}
\includegraphics[width=\textwidth]{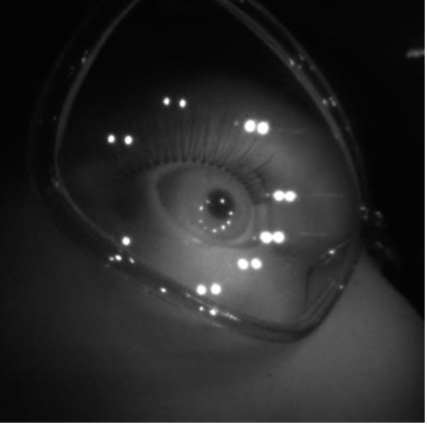}
\end{subfigure}
\,
\begin{subfigure}{0.15\columnwidth}
\includegraphics[width=\textwidth]{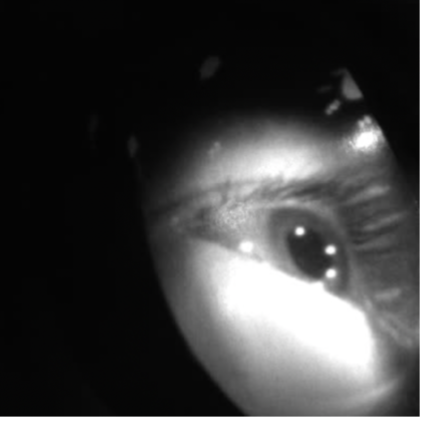}
\end{subfigure}
\caption{Samples of PA recordings from \tdataset. The top row represents the PAI captured in RGB (visible) spectrum, while
middle and bottom rows depict the NIR recordings without and with glasses, respectively, as acquired by the internal (right) camera of the \tdevice. 
From left to right, each column presents a sample of the type of PAIs belonging to the attack series from 2--7.}
\label{fig:pa_samples}
\end{figure}

%% file: sections/use_cases.tex
%

This section outlines the potential use-cases of the newly created \tdataset
dataset, emphasizing its applicability in biometric applications such as iris
and periocular detection/recognition, presentation attack detection (PAD), as
well as intriguing tasks (or sub-tasks) associated with biometrics such as
semantic segmentation focusing on the eye, iris, and sclera regions. The
overview of these use-cases is illustrated in Fig.~\ref{fig:use_cases}

The use cases discussed here, although well-established in research and
development, are highlighted more as possibilities than substantiated claims of
direct applicability. The \tdataset dataset can serve as a benchmark to evaluate
the effectiveness of current HMD data for these applications. Alternatively, it
can be instrumental in developing (or fine-tuning) applications specifically for
data obtained from HMD devices. The present data can be characterised by
non-frontal views of periocular regions, relatively low spatial resolutions, and
limited resources. Thus, some applications discussed in this section may prove
to be quite challenging. This, however, presents interesting research problems
to transform this data into better quality as well as to develop robust
applications.

Needless to mention, the advances in HMD technology will not only enhance the
quality of the data but also expand the scope of applications.

\subsection{Iris and Periocular Recognition}

The \tdataset dataset comprises 900 video recordings from 25 \tbf subjects.
Each subject participated in sessions combining three gaze and two eye-wear
variations, thus providing a valuable resource for sampling and analyzing iris
and periocular regions for detection and recognition tasks. When sampled at the
frame level, videos with moving gaze can be treated as individual frames with
varied gazes. These individual frames facilitate the localization and enrollment
of the eye region, making them suitable for both detection and recognition
purposes. Detection involves localizing specific regions of interest (RoI), such
as the iris, eye, or periocular regions, while recognition can be performed in
both \textit{1:1} verification and \textit{1:N} identification scenarios.

Based on the design of the pipeline, recent iris/periocular recognition methods
can be categorized into feature-extraction based or classification based.
Feature-extraction based methods employ models, often deep convolutional neural
network (CNN) architectures, as feature extractors to obtain a compact
representation (\textit{embedding}) of a preprocessed input, (\textit{i.e.},
iris/periocular image). In these methods, preprocessing and matching/scoring are
typically separated from feature extraction. Classification based methods, on
the other hand, treat the recognition problem in an end-to-end manner, training
the iris dataset in a supervised learning setting.

An overview of iris recognition methods using handcrafted features and classical
machine learning techniques can be found in \cite{bowyer2016handbook}. A recent
work by Nguyen \textit{et al.} provides a systematic survey of deep learning
(DL)-based methods for iris recognition \cite{nguyen2024deep}. Another
comprehensive review of iris recognition methods is presented in
\cite{yin2023deep}, which details methods from both categories and various
preprocessing techniques for the first category (non end-to-end methods).
Alonso-Fernandez and Bigun discuss several handcrafted feature-based methods for
periocular recognition, categorizing them into texture, shape, and color-based
methods \cite{alonso2016survey}.

Iris recognition remains a topic of significant interest within the biometric
community. Over the past decade, several challenges have been organized to
advance iris recognition performance \cite{zhang2014first, zhang2016btas}. The
challenge described in \cite{zhang2016btas} was specifically dedicated to
mobile iris recognition. Boutros \textit{et al.} benchmarked the
OpenEDS~\cite{garbin2019openeds}- an HMD dataset-- for iris
recognition~\cite{9107939, boutros2020benchmarking}. Although this dataset has
been acquired from an HMD device, it captures nearly frontal view of the eye
regions making it easy to detect, localize, and recognize an individual.

The use of HMD data for biometric recognition is still in its early stages and
is yet to gain mainstream attention. A major challenge posed by HMD-based iris
detection is the oblique view of the RoI, which often results in localization
failures and leads to failure to acquire (FtA). The non-frontal view, even
after normalization, may cause feature/data distortion. This, combined with the
relatively small size of the RoI, results in the loss of fine, subtle features
that may present discriminatory information. Several quality metrics for iris
and periocular images, including low-quality images, have been presented in
\cite{bowyer2016handbook}. A study of the quality assessment of HMD-based data
for recognition purposes may provide useful insights towards designing new
methods.


\subsection{Iris and Periocular PAD}

For a biometric authentication system to be practically deployable, its
resistance to presentation attacks (PAs) is crucial. The importance of PAD for
iris and periocular traits has been well recognized by the biometric community.
Several works by A. Czajka, K. Bowyer, and colleagues present comprehensive
reviews of iris PAD methods and
datasets~\cite{czajka2018presentation,boyd2020iris}. However, these works do not
discuss datasets or PAD methods for HMD-based datasets, likely due to a lack of
relevant data. With \tdataset, we provide the first PAD dataset acquired using
HMD devices, containing approximately 1100 short videos of PAs. Through a
combination of attack instruments for the eyes and surrounding regions (refer to
Section~\ref{sec:dataset}), we have created a variety of attack scenarios for
PAD tasks. The inclusion of print attacks of the \tbf samples allows for the
vulnerability analysis of different PAD methods.

Several studies have also proposed approaches that combine both iris and
periocular regions for PA detection \cite{raja2016presentation,
hoffman2019iris}. Such approaches can be explored for the \tdataset dataset, if
the detection and localization of iris does not yield satisfactory results.

LivDet-Iris\footnote{\url{https://livdet.org}} is a well-known competition
regularly organized to compare liveness detection methods (\textit{i.e.}, to
differentiate between real human and fake samples). The fifth installment of
this competition was held in conjunction with IJCB in 2023
\cite{tinsley2023iris}. Despite the participation of reputable academic
institutions, the average classification error rates were as high as 22--37\%
(the variation refers to the weighing mechanism of different PAIs). Although the
quality of data and types of attacks in the competition datasets and \tdataset
dataset differ, this highlights the challenging and unresolved nature of the
iris and/or periocular PAD problem.

\subsection{Semantic Segmentation}

Semantic segmentation of the eye to localize components such as the iris and
sclera has demonstrated its usefulness in both biometric and non-biometric
tasks. This segmentation process is often an integral crucial stage in
recognition and PAD-related tasks discussed in the previous section. The
creators of the OpenEDS dataset have also highlighted segmentation as a key
application of interest \cite{garbin2019openeds}. Segmentation of individual
traits, such as the iris \cite{adegoke2013iris} and sclera \cite{rot2020deep},
has received considerable attention in the literature. Recently, multi-class
semantic segmentation--which involves the simultaneous detection and
localization of various parts of the eye region-- has gained interest as
well~\cite{chaudhary2019ritnet, perry2019minenet, valenzuela2020towards}. 

Similar to the previous use-cases, the topic of eye segmentation has been
further advanced by numerous competitions. For instance, challenges such as
NIR-ISL~\cite{wang2021nir} and SSRBC~\cite{das2023sclera} focus specifically on
iris and sclera segmentation, respectively. The \tdataset dataset (after
annotations) can also be used to benchmark the performance of existing
segmentation methods, with reference to biometric applications, on realistic HMD
data captured in the NIR spectrum. Given the variation in the exact angle of
capture due to differences in subjects' face shapes and inter-eye distances, the
frames in the \tdataset dataset present a challenging dataset. As typical
segmentation methods are usually trained on frontal views of the eye, one may be
required to address this domain gap by designing implicit or explicit affine
transformations.

%
%

%% file: sections/conc.tex
In this work, we introduced the \tdataset dataset, a novel dataset of periocular
videos captured using the VR device, \tdevice. This dataset is the first
publicly available resource offering realistic, non-frontal views of the
periocular region, comprising 900 short videos from 25 subjects and 1104
presentation attack (PA) videos using 92 different attack instruments. The
dataset includes diverse and challenging scenarios such as steady gaze, moving
gaze, and partially closed eyes; with an equal split of recordings with and
without glasses.

The HMD-captured data are relatively novel and present significant potential for
various applications that are yet to be fully explored. However, these also pose
unique challenges such as non-frontal views, user-specific fitment issues, and
low resolutions. With the \tdataset dataset, we hope to provide a valuable
resource for understanding and advancing biometric applications associated with
AR and VR devices. The inclusion of various attack recordings also provides an
opportunity to enhance \textit{anti-spoofing} measures for VR-based
authentication systems.

We have outlined the data collection process and associated challenges,
providing useful insights for researchers aiming to acquire similar data from VR
devices. The \tdataset dataset is publicly available to support advancements in
biometric research, including authentication, semantic segmentation, and
presentation attack detection (PAD), along with related sub-tasks.

%% file: main.bbl
\begin{thebibliography}{10}

\bibitem{radianti2020systematic}
Jaziar Radianti, Tim~A Majchrzak, Jennifer Fromm, and Isabell Wohlgenannt,
\newblock ``A systematic review of immersive virtual reality applications for
  higher education: Design elements, lessons learned, and research agenda,''
\newblock {\em Computers \& education}, vol. 147, pp. 103778, 2020.

\bibitem{kaminska2019virtual}
Dorota Kami{\'n}ska et~al.,
\newblock ``Virtual reality and its applications in education: Survey,''
\newblock {\em Information}, vol. 10, no. 10, pp. 318, 2019.

\bibitem{xie2021review}
Biao Xie et~al.,
\newblock ``A review on virtual reality skill training applications,''
\newblock {\em Frontiers in Virtual Reality}, vol. 2, pp. 645153, 2021.

\bibitem{chiang2022augmented}
Feng-Kuang Chiang, Xiaojing Shang, and Lu~Qiao,
\newblock ``Augmented reality in vocational training: A systematic review of
  research and applications,''
\newblock {\em Computers in Human Behavior}, vol. 129, pp. 107125, 2022.

\bibitem{yeung2021virtual}
Andy Wai~Kan Yeung et~al.,
\newblock ``Virtual and augmented reality applications in medicine: analysis of
  the scientific literature,''
\newblock {\em Journal of medical internet research}, vol. 23, no. 2, pp.
  e25499, 2021.

\bibitem{javaid2020virtual}
Mohd Javaid and Abid Haleem,
\newblock ``Virtual reality applications toward medical field,''
\newblock {\em Clinical Epidemiology and Global Health}, vol. 8, no. 2, pp.
  600--605, 2020.

\bibitem{suh2023current}
Irene Suh, Tess McKinney, and Ka-Chun Siu,
\newblock ``Current perspective of metaverse application in medical education,
  research and patient care,''
\newblock in {\em Virtual Worlds}. MDPI, 2023, vol.~2, pp. 115--128.

\bibitem{boutros2020iris}
Fadi Boutros, Naser Damer, Kiran Raja, Raghavendra Ramachandra, Florian
  Kirchbuchner, and Arjan Kuijper,
\newblock ``Iris and periocular biometrics for head mounted displays:
  Segmentation, recognition, and synthetic data generation,''
\newblock {\em Image and Vision Computing}, vol. 104, pp. 104007, 2020.

\bibitem{boutros2020benchmarking}
Fadi Boutros, Naser Damer, Kiran Raja, Raghavendra Ramachandra, Florian
  Kirchbuchner, and Arjan Kuijper,
\newblock ``On benchmarking iris recognition within a head-mounted display for
  ar/vr applications,''
\newblock in {\em Proceedings of the IEEE International Joint Conference on
  Biometrics (IJCB)}. IEEE, 2020, pp. 1--10.

\bibitem{morales2019introduction}
Aythami Morales, Julian Fierrez, Javier Galbally, and Marta Gomez-Barrero,
\newblock ``Introduction to iris presentation attack detection,''
\newblock {\em Handbook of Biometric Anti-Spoofing: Presentation Attack
  Detection}, pp. 135--150, 2019.

\bibitem{czajka2018presentation}
Adam Czajka and Kevin~W Bowyer,
\newblock ``Presentation attack detection for iris recognition: An assessment
  of the state-of-the-art,''
\newblock {\em ACM Computing Surveys (CSUR)}, vol. 51, no. 4, pp. 1--35, 2018.

\bibitem{wang2021edge}
Zhimin Wang, Yuxin Zhao, Yunfei Liu, and Feng Lu,
\newblock ``Edge-guided near-eye image analysis for head mounted displays,''
\newblock in {\em Proceedings of IEEE International Symposium on Mixed and
  Augmented Reality (ISMAR)}. IEEE, 2021, pp. 11--20.

\bibitem{clay2019eye}
Viviane Clay, Peter K{\"o}nig, and Sabine Koenig,
\newblock ``Eye tracking in virtual reality,''
\newblock {\em Journal of eye movement research}, vol. 12, no. 1, 2019.

\bibitem{kapp2021arett}
Sebastian Kapp, Michael Barz, Sergey Mukhametov, Daniel Sonntag, and Jochen
  Kuhn,
\newblock ``{ARETT: augmented reality eye tracking toolkit for head mounted
  displays},''
\newblock {\em Sensors}, vol. 21, no. 6, pp. 2234, 2021.

\bibitem{bozkir2023eye}
Efe Bozkir et~al.,
\newblock ``Eye-tracked virtual reality: A comprehensive survey on methods and
  privacy challenges,''
\newblock {\em arXiv preprint arXiv:2305.14080}, 2023.

\bibitem{9107939}
Fadi Boutros, Naser Damer, Kiran Raja, Raghavendra Ramachandra, Florian
  Kirchbuchner, and Arjan Kuijper,
\newblock ``Periocular biometrics in head-mounted displays: A sample selection
  approach for better recognition,''
\newblock in {\em International Workshop on Biometrics and Forensics (IWBF)},
  2020, pp. 1--6.

\bibitem{wang2021nir}
Caiyong Wang et~al.,
\newblock ``Nir iris challenge evaluation in non-cooperative environments:
  Segmentation and localization,''
\newblock in {\em 2021 IEEE International joint conference on biometrics
  (IJCB)}. IEEE, 2021, pp. 1--10.

\bibitem{hoffman2019iris}
Steven Hoffman, Renu Sharma, and Arun Ross,
\newblock ``{Iris + Ocular: Generalized iris presentation attack detection
  using multiple convolutional neural networks},''
\newblock in {\em International Conference on Biometrics (ICB)}. IEEE, 2019,
  pp. 1--8.

\bibitem{casia_iris}
{Chinese Academy of Sciences},
\newblock ``Casia iris image database,'' 2002,
\newblock http://biometrics.idealtest.org.

\bibitem{zanlorensi2022ocular}
Luiz~A Zanlorensi, Rayson Laroca, Eduardo Luz, Alceu~S Britto~Jr, Luiz~S
  Oliveira, and David Menotti,
\newblock ``Ocular recognition databases and competitions: A survey,''
\newblock {\em Artificial Intelligence Review}, vol. 55, no. 1, pp. 129--180,
  2022.

\bibitem{omelina2021survey}
Lubos Omelina, Jozef Goga, Jarmila Pavlovicova, Milos Oravec, and Bart Jansen,
\newblock ``A survey of iris datasets,''
\newblock {\em Image and Vision Computing}, vol. 108, pp. 104109, 2021.

\bibitem{yambay2017livdet}
David Yambay et~al.,
\newblock ``Livdet iris 2017--iris liveness detection competition 2017,''
\newblock in {\em IEEE International Joint Conference on Biometrics (IJCB)}.
  IEEE, 2017, pp. 733--741.

\bibitem{tinsley2023iris}
Patrick Tinsley et~al.,
\newblock ``Iris liveness detection competition (livdet-iris)--the 2023
  edition,''
\newblock in {\em IEEE International Joint Conference on Biometrics (IJCB)}.
  IEEE, 2023, pp. 1--10.

\bibitem{proencca2005ubiris}
Hugo Proen{\c{c}}a and Lu{\'\i}s~A Alexandre,
\newblock ``Ubiris: A noisy iris image database,''
\newblock in {\em International Conference on Image Analysis and Processing
  (ICIAP)}. Springer, 2005, pp. 970--977.

\bibitem{proencca2009ubiris}
Hugo Proen{\c{c}}a, Silvio Filipe, Ricardo Santos, Joao Oliveira, and Luis~A
  Alexandre,
\newblock ``The ubiris. v2: A database of visible wavelength iris images
  captured on-the-move and at-a-distance,''
\newblock {\em IEEE Transactions on Pattern Analysis and Machine Intelligence},
  vol. 32, no. 8, pp. 1529--1535, 2009.

\bibitem{teo2010robust}
Chuan~Chin Teo, Han~Foon Neo, GKO Michael, Connie Tee, and KS~Sim,
\newblock ``A robust iris segmentation with fuzzy supports,''
\newblock in {\em International Conference on Neural Information Processing
  (ICONIP)}. Springer, 2010, pp. 532--539.

\bibitem{nguyen2024deep}
Kien Nguyen, Hugo Proen{\c{c}}a, and Fernando Alonso-Fernandez,
\newblock ``Deep learning for iris recognition: A survey,''
\newblock {\em ACM Computing Surveys}, vol. 56, no. 9, pp. 1--35, 2024.

\bibitem{mcmurrough2013dataset}
Christopher~D McMurrough, Vangelis Metsis, Dimitrios Kosmopoulos, Ilias
  Maglogiannis, and Fillia Makedon,
\newblock ``A dataset for point of gaze detection using head poses and eye
  images,''
\newblock {\em Journal on Multimodal User Interfaces}, vol. 7, pp. 207--215,
  2013.

\bibitem{tonsen2016labelled}
Marc Tonsen, Xucong Zhang, Yusuke Sugano, and Andreas Bulling,
\newblock ``Labelled pupils in the wild: a dataset for studying pupil detection
  in unconstrained environments,''
\newblock in {\em Proceedings of the ninth biennial ACM symposium on eye
  tracking research \& applications}, 2016, pp. 139--142.

\bibitem{kim2019nvgaze}
Joohwan Kim et~al.,
\newblock ``Nvgaze: An anatomically-informed dataset for low-latency, near-eye
  gaze estimation,''
\newblock in {\em Proceedings of the CHI conference on human factors in
  computing systems}, 2019, pp. 1--12.

\bibitem{garbin2019openeds}
Stephan~J Garbin, Yiru Shen, Immo Schuetz, Robert Cavin, Gregory Hughes, and
  Sachin~S Talathi,
\newblock ``Openeds: Open eye dataset,''
\newblock {\em arXiv preprint arXiv:1905.03702}, 2019.

\bibitem{palmero2020openeds2020}
Cristina Palmero, Abhishek Sharma, Karsten Behrendt, Kapil Krishnakumar, Oleg~V
  Komogortsev, and Sachin~S Talathi,
\newblock ``Openeds2020: Open eyes dataset,''
\newblock {\em arXiv preprint arXiv:2005.03876}, 2020.

\bibitem{bowyer2016handbook}
Kevin~W Bowyer and Mark~J Burge,
\newblock {\em Handbook of iris recognition},
\newblock Springer, 2016.

\bibitem{yin2023deep}
Yimin Yin, Siliang He, Renye Zhang, Hongli Chang, Xu~Han, and Jinghua Zhang,
\newblock ``Deep learning for iris recognition: a review,''
\newblock {\em arXiv preprint arXiv:2303.08514}, 2023.

\bibitem{alonso2016survey}
Fernando Alonso-Fernandez and Josef Bigun,
\newblock ``A survey on periocular biometrics research,''
\newblock {\em Pattern Recognition Letters}, vol. 82, pp. 92--105, 2016.

\bibitem{zhang2014first}
Man Zhang et~al.,
\newblock ``{The first ICB* competition on iris recognition},''
\newblock in {\em IEEE International Joint Conference on Biometrics}. IEEE,
  2014, pp. 1--6.

\bibitem{zhang2016btas}
Man Zhang, Qi~Zhang, Zhenan Sun, Shujuan Zhou, and Nasir~Uddin Ahmed,
\newblock ``The btas competition on mobile iris recognition,''
\newblock in {\em 8th international conference on biometrics theory,
  applications and systems (BTAS)}. IEEE, 2016, pp. 1--7.

\bibitem{boyd2020iris}
Aidan Boyd, Zhaoyuan Fang, Adam Czajka, and Kevin~W Bowyer,
\newblock ``Iris presentation attack detection: Where are we now?,''
\newblock {\em Pattern Recognition Letters}, vol. 138, pp. 483--489, 2020.

\bibitem{raja2016presentation}
Kiran~B Raja, R~Raghavendra, Jean-Noel Braun, and Christoph Busch,
\newblock ``Presentation attack detection using a generalizable statistical
  approach for periocular and iris systems,''
\newblock in {\em Proceedings of the International Conference of the Biometrics
  Special Interest Group (BIOSIG)}, 2016, pp. 1--6.

\bibitem{adegoke2013iris}
BO~Adegoke, EO~Omidiora, SA~Falohun, and JA~Ojo,
\newblock ``Iris segmentation: a survey,''
\newblock {\em International Journal of Modern Engineering Research (IJMER)},
  vol. 3, no. 4, pp. 1885--1889, 2013.

\bibitem{rot2020deep}
Peter Rot, Matej Vitek, Klemen Grm, {\v{Z}}iga Emer{\v{s}}i{\v{c}}, Peter Peer,
  and Vitomir {\v{S}}truc,
\newblock ``Deep sclera segmentation and recognition,''
\newblock {\em Handbook of vascular biometrics}, pp. 395--432, 2020.

\bibitem{chaudhary2019ritnet}
Aayush~K Chaudhary, Rakshit Kothari, Manoj Acharya, Shusil Dangi, Nitinraj
  Nair, Reynold Bailey, Christopher Kanan, Gabriel Diaz, and Jeff~B Pelz,
\newblock ``Ritnet: Real-time semantic segmentation of the eye for gaze
  tracking,''
\newblock in {\em IEEE/CVF International Conference on Computer Vision Workshop
  (ICCVW)}. IEEE, 2019, pp. 3698--3702.

\bibitem{perry2019minenet}
Jonathan Perry and Amanda Fernandez,
\newblock ``Minenet: A dilated cnn for semantic segmentation of eye features,''
\newblock in {\em Proceedings of the IEEE/CVF international conference on
  computer vision workshops}, 2019, pp. 0--0.

\bibitem{valenzuela2020towards}
Andres Valenzuela, Claudia Arellano, and Juan~E Tapia,
\newblock ``Towards an efficient segmentation algorithm for near-infrared eyes
  images,''
\newblock {\em IEEE Access}, vol. 8, pp. 171598--171607, 2020.

\bibitem{das2023sclera}
Abhijit Das et~al.,
\newblock ``Sclera segmentation and joint recognition benchmarking competition:
  Ssrbc 2023,''
\newblock in {\em IEEE International Joint Conference on Biometrics (IJCB)}.
  IEEE, 2023, pp. 1--10.

\end{thebibliography}
